\documentclass{article}

    \usepackage[final]{iai_neurips_2024}

\usepackage[utf8]{inputenc} % allow utf-8 input
\usepackage[T1]{fontenc}    % use 8-bit T1 fonts
\usepackage{hyperref}       % hyperlinks
\usepackage{url}            % simple URL typesetting
\usepackage{booktabs}       % professional-quality tables
\usepackage{amsfonts}       % blackboard math symbols
\usepackage{nicefrac}       % compact symbols for 1/2, etc.
\usepackage{microtype}      % microtypography
\usepackage{xcolor}         % colors

\usepackage{algorithm}
\usepackage{algorithmic}
\usepackage{aligned-overset}
\usepackage{amsmath}
\usepackage{amssymb}
\usepackage{amsthm}
\usepackage{mathtools}

\usepackage{graphicx}
\usepackage{subcaption}

\title{Disentangling Mean Embeddings for Better Diagnostics of Image Generators}

\author{%
  Sebastian G.~Gruber \\
  German Cancer Consortium (DKTK), partner site Frankfurt/Mainz, \\ a partnership between DKFZ and UCT Frankfurt-Marburg, Germany, Frankfurt am Main, Germany \\ German Cancer Research Center (DKFZ), Heidelberg, Germany \\ Goethe University Frankfurt, Germany \\
  \texttt{sebastian.gruber@dkfz.de} \\
  \And
  Pascal Tobias~Ziegler \\
  Goethe University Frankfurt, Germany \\
  \AND
  Florian Buettner \\
  German Cancer Consortium (DKTK), partner site Frankfurt/Mainz, \\ a partnership between DKFZ and UCT Frankfurt-Marburg, Germany, Frankfurt am Main, Germany \\ German Cancer Research Center (DKFZ), Heidelberg, Germany \\   Frankfurt Cancer Institute (FCI), Germany \\
  Goethe University Frankfurt, Germany \\
}

\newtheorem{theorem}{Theorem}
\newtheorem{corollary}{Corollary}

\begin{document}

\maketitle

\begin{abstract}
The evaluation of image generators remains a challenge due to the limitations of traditional metrics in providing nuanced insights into specific image regions.
This is a critical problem as not all regions of an image may be learned with similar ease.
In this work, we propose a novel approach to disentangle the cosine similarity of mean embeddings into the product of cosine similarities for individual pixel clusters via central kernel alignment.
Consequently, we can quantify the contribution of the cluster-wise performance to the overall image generation performance.
We demonstrate how this enhances the explainability and the likelihood of identifying pixel regions of model misbehavior across various real-world use cases.
\end{abstract}

\section{Introduction}
\label{sec:intro}

The increasing prevalence of Artificial Intelligence (AI), particularly with the rise of sophisticated generative models like image generators, has brought about a transformative shift beyond the field of machine learning \citep{singh2021medical, mirsky2021creation, oppenlaender2022creativity}.
However, the evaluation of the outputs from these models, especially in the realm of image generation, continues to pose a significant challenge \citep{benny2021evaluation, elasri2022image, xu2024imagereward}.
Traditional evaluation metrics, like the maximum mean discrepancy (MMD) \citep{JMLR:v13:gretton12a}, the Inception Score Criterion (ISC) \citep{salimans2016improved}, the Fréchet Inception Distance (FID) \citep{heusel2017gans}, or the Kernel Inception Distance (KID) \citep{binkowski2018demystifying}, fall short in providing a nuanced understanding of specific image regions, thereby limiting their effectiveness in assessing model performance comprehensively.
The MMD is the squared distance between the mean embeddings of two distributions.
A mean embedding is the expectation of a kernel-induced feature map based on the respective distribution.
As we will see, we can decompose, which we refer to as \emph{disentangle} based on \citep{vedral2002role}, mean embeddings as a tensor product under certain conditions.
Depending on the kernel choice, the MMD can be used without an external model, but it cannot be decomposed in a meaningful way even when disentangled mean embeddings exist.
This disallows more fine-grained interpretations of model performance by current evaluation approaches.
In the context of this work, interpretation, interpretability, and explainability refer to performance and error assignment to different image regions, which increases human oversight and understanding, contrary to ``black box'' evaluation approaches only assessing entire images \citep{castelvecchi2016can, phillips2021four, longo2024explainable}.

In the present work, we \textbf{contribute} a novel approach based on disentangling the mean embedding of an image space into mean embeddings of independent clusters of pixels.
Further, we show that the cosine similarity of the mean embeddings can be disentangled into the product of the cosine similarities for each respective cluster.
This enables the evaluation and interpretation of the model performance of each cluster in isolation, significantly enhancing diagnostics of image generation and the likelihood of identifying the source pixel region of model misbehavior.
We illustrate the gain in interpretability by monitoring the generalization performance of DCGAN \citep{radford2015unsupervised} and DDPM \citep{ho2020denoising} architectures trained on CelebA \citep{liu2015deep} and ChestMNIST \citep{medmnistv1} datasets.
%The findings of this work contribute to a deeper comprehension and improved evaluation of generative models, thereby broadening a reliable application across various domains.

\section{Preliminaries on Mean Embeddings}
\label{sec:preliminaries}

In this section, we introduce the necessary background for mean embeddings and central kernel alignment.
Given a nonempty set $\mathcal{X}$ and a positive semi-definite (p.s.d.) kernel $k \colon \mathcal{X} \times \mathcal{X} \to \mathbb{R}$, there exists $\phi$ and an inner product $\left\langle . , . \right\rangle$ such that $k \left( x, y \right) = \left\langle \phi \left( x \right), \phi \left( y \right) \right\rangle$ \citep{scholkopf2002learning}.
The associated RKHS is defined as the completion $\mathcal{H} = \overline{\operatorname{span} \left\{ \phi \left( x \right) \mid x \in \mathcal{X} \right\}}$, where $\left\langle . , . \right\rangle_{\mathcal{H}} \coloneqq \left\langle . , . \right\rangle$ and $\left\lVert h \right\rVert_{\mathcal{H}} \coloneqq \sqrt{\left\langle h, h \right\rangle_{\mathcal{H}}}$ for $h \in \mathcal{H}$.
For a distribution $P$ with support $\mathcal{X}$ the \textbf{mean embedding} in $\mathcal{H}$ is defined via
\begin{equation}
    \mu_P \coloneqq \mathbb{E}_{X \sim P} \left[ \phi \left( X \right) \right].
\end{equation}
Given another distribution $Q$ with similar support, \citet{JMLR:v13:gretton12a} introduce the \textbf{maximum mean discrepancy} as the squared distance between mean embeddings $\mu_P$ and $\mu_Q$ defined by
\begin{equation}
    \operatorname{MMD}_k^2 \left(P, Q \right) \coloneqq \left\lVert \mu_P - \mu_Q \right\rVert_{\mathcal{H}}^2.
\end{equation}
If $k$ is a characteristic kernel, i.e. $\mu_{(.)}$ is an injective function for a set of distributions including $P$ and $Q$, then it holds $P = Q \iff \operatorname{MMD}_k^2 \left( P, Q \right) = 0$ \citep{JMLR:v13:gretton12a}.
However, for the central research question of this paper, we will discover that the MMD cannot be disentangled into the MMD values of different input regions.
Fortunately, there also exist other approaches to quantify the similarity of vectors $\mu_P$ and $\mu_Q$ in the RKHS $\mathcal{H}$.
One of the most classical metrics is the cosine similarity defined between vectors $v,w \in \mathbb{R}^d$ via $\operatorname{cos} \left( v, w \right) = \frac{\left\langle v, w \right\rangle}{\left\lVert v \right\rVert \left\lVert w \right\rVert}$.
Compared to the squared distance, it has the benefit of being easier to interpret as it lies within $[-1, 1]$ with $v=w \implies \operatorname{cos} \left( v, w \right) = 1$.
For general RKHS, the cosine similarity was already studied implicitly as the inner product of quantum mean embeddings in \citep{kubler2019quantum}.
To receive an analogous definition to the MMD, we define the cosine similarity between the mean embeddings $\mu_P$ and $\mu_Q$ as the \textbf{cosine mean similarity} given by
\begin{equation}
    \operatorname{CMS}_k \left(P, Q \right) \coloneqq \frac{\left\langle \mu_P, \mu_Q \right\rangle_{\mathcal{H}}}{\left\lVert \mu_P \right\rVert_{\mathcal{H}} \left\lVert \mu_Q \right\rVert_{\mathcal{H}}}.
\end{equation}
While the $\operatorname{MMD}$ can be seen as the generalization of the squared distance to possibly infinite dimensional RKHS, the $\operatorname{CMS}$ is the analogous generalization of the cosine similarity.
If $k$ is a $c_0$ universal kernel and $P,Q$ are Borel probability measures, then it holds $P = Q \iff \operatorname{CMS}_k \left( P, Q \right) = 1$ \citep{kubler2019quantum}.
If $k$ is $c_0$ universal, then it follows that it is also characteristic, but the opposite does not always hold \citep{sriperumbudur2011universality}.
Consequently, $\operatorname{CMS}$ and $\operatorname{MMD}$ only share the uniqueness of their optimum for $c_0$ universal kernels.
Examples of $c_0$ universal kernels are the RBF kernel $k_{\mathrm{rbf}} \left( x, y \right) = \exp \left( -\gamma \left\lVert x - y \right\rVert^2_2 \right)$, the Laplacian kernel, the Matérn kernel, or any other characteristic and translation invariant kernel \citep{sriperumbudur2011universality}.
Given datasets $\mathbf{X} = \left( X_1, \dots, X_n \right) \overset{iid}{\sim} P$ and $\mathbf{Y} = \left( Y_1, \dots, Y_m \right) \overset{iid}{\sim} Q$ of independently and identically distributed (iid) random variables, an empirical estimator for $\operatorname{CMS}_k \left( P, Q \right)$ can be defined via
\begin{equation}
\label{eq:CMS_est}
    \widehat{\operatorname{CMS}}_k \left( \mathbf{X}, \mathbf{Y} \right) \coloneqq \frac{\sum_{i=1}^n \sum_{j=1}^m k \left( X_i, Y_j \right)}{\sqrt{\sum_{i=1}^n \sum_{j=1}^n k \left( X_i, X_j \right)}\sqrt{\sum_{i=1}^m \sum_{j=1}^m k \left( Y_i, Y_j \right)}}.
\end{equation}

Using kernels with images in practice usually follows either of two approaches: Either the image is flattened and each pixel is an entry in the vector provided as argument for the kernel \citep{scholkopf1997support, gruber2024biasvariancecovariance}, or the image is encoded into a smaller-dimensional semantic vector space, which is then the argument for the kernel \citep{binkowski2018demystifying}.
While the latter approach gained more prominence in recent years \citep{benny2021evaluation, xu2024imagereward}, the encoding is usually learned by a neural network and not interpretable.
In the following of this work, we focus on the former approach, which allows to disentangle the image provided that the chosen kernel is a product of pixel-wise kernels.
Specifically, we assume that for flattened images $x = \left( x_1, \dots, x_d \right)^\intercal \in \mathbb{R}^d$ and $y = \left( y_1, \dots, y_d \right) \in \mathbb{R}^d$ with $d$ pixels the image-wise kernel $k_{\mathrm{img}} \colon \mathcal{X}_{\mathrm{img}} \times \mathcal{X}_{\mathrm{img}} \to \mathbb{R}$ with $\mathcal{X}_{\mathrm{img}} \subseteq \mathbb{R}^d$ can be decomposed into a product $k_{\mathrm{img}} \left( x, y \right) = k_{\mathrm{pxl}}^{\otimes d} \left( \left( x_1, \dots, x_d \right), \left( y_1, \dots, y_d \right) \right) = k_{\mathrm{pxl}} \left( x_1, y_1 \right) \cdots k_{\mathrm{pxl}} \left( x_d, y_d \right)$ for a pixel-wise kernel $k_{\mathrm{pxl}} \colon \mathcal{X}_{\mathrm{pxl}} \times \mathcal{X}_{\mathrm{pxl}} \to \mathbb{R}$ with $\mathcal{X}_{\mathrm{pxl}} \subseteq \mathbb{R}$.
Note that this assumes grey-scale images for simplicity, however colored images (with three color channels) can be easily represented 
by assuming $\mathcal{X}_{\mathrm{img}} \subseteq \mathbb{R}^{3d}$ and $\mathcal{X}_{\mathrm{pxl}} \subseteq \mathbb{R}^3$.
The RBF and Laplacian kernels are such product kernels, since we can write $k_{\mathrm{rbf}} \left( x, y \right) = \exp \left( -\gamma \left(x_1 - y_1 \right)^2 \right) \cdots \exp \left( -\gamma \left(x_d - y_d \right)^2 \right)$. %and $k_{\mathrm{lap}} \left( x, y \right) = \exp \left( -\gamma \left\lvert x_1 - y_1 \right\rvert \right) \cdots \exp \left( -\gamma \left\lvert x_d - y_d \right\rvert \right)$.
In general, these are special cases of product kernels discussed in \citep{szabo2018characteristic}.
The image-wise RKHS $\mathcal{H}_{\operatorname{img}}$ and feature map $\phi_{\operatorname{img}} \colon \mathcal{X}_{\mathrm{img}} \to \mathcal{H}_{\operatorname{img}}$ associated with $k_{\mathrm{img}}$ can also be decomposed into a tensor product space $\mathcal{H}_{\operatorname{img}} = \underbrace{\mathcal{H}_{\operatorname{pxl}} \otimes \dots \otimes \mathcal{H}_{\operatorname{pxl}}}_{d \text{ times}}$ of the pixel-wise RKHS $\mathcal{H}_{\operatorname{pxl}}$ and a tensor product
\begin{equation}
\begin{split}
    \phi_{\operatorname{img}} \left( x \right) & = {\left( \phi_{\operatorname{pxl}} \otimes \dots \otimes \phi_{\operatorname{pxl}} \right)}_{} \left( x_1, \dots, x_d \right) = \bigotimes_{i=1}^d \phi_{\operatorname{pxl}} \left( x_i \right)    
\end{split}
\end{equation}
of the pixel-wise feature maps $\phi_{\operatorname{pxl}} \colon \mathcal{X}_{\mathrm{pxl}} \to \mathcal{H}_{\operatorname{pxl}}$ associated with $k_{\mathrm{pxl}}$ \citep{szabo2018characteristic}.
However, when we compute the respective mean embedding for a distribution $P_{\mathrm{img}}$ of a random image $X = \left(X_1, \dots, X_d \right)$, then we cannot decompose it in general into pixel-wise mean embeddings since
\begin{equation}
    \mu_{P_{\mathrm{img}}} = \mathbb{E}_{X \sim P_{\mathrm{img}}} \left[ \phi_{\operatorname{img}} \left( X \right) \right] = \mathbb{E}_{X \sim P_{\mathrm{img}}} \left[ \bigotimes_{i=1}^d \phi_{\operatorname{pxl}} \left( X_i \right) \right] \neq  \bigotimes_{i=1}^d \mathbb{E}_{X_i \sim P_i} \left[ \phi_{\operatorname{pxl}} \left( X_i \right) \right]
\label{eq:perfect_disent}
\end{equation}
where $P_i$ are the marginal distributions of pixel indices $i=1 \dots d$.
Such a pixel-wise decomposition, which we also refer to as disentanglement, is usually not possible in practice since pixels are correlated.
However, for interpretability purposes, we do not necessarily require a complete disentanglement of all individual pixels, but it may suffice to discover disentangled clusters of pixels.
To quantify to what degree this is possible in practice, we require the following. \\
The cross-covariance matrix between an $\mathbb{R}^d$-valued random variable $X$ and an $\mathbb{R}^{d^\prime}$-valued random variable $Y$ is defined by 
\begin{equation}
    \operatorname{Cov} \left( X, Y \right) \coloneqq \mathbb{E} \left[ \left( X - \mathbb{E} \left[ X \right] \right)\left( Y - \mathbb{E} \left[ Y \right] \right)^\intercal \right] \in \mathbb{R}^{d \times d^\prime}.
\end{equation}
Any matrix in $\mathbb{R}^{d \times d^\prime}$ may also be seen as a linear operator from $\mathbb{R}^{d^\prime} \to \mathbb{R}^{d}$.
Consequently, given another p.s.d. kernel $k^\prime$ with RKHS $\mathcal{H}^\prime$, a generalization of the cross-covariance matrix to $\mathcal{H}$-valued and $\mathcal{H}^\prime$-valued random variables $\phi \left( X \right)$ and $\phi^\prime \left( Y \right)$ is given via the cross-covariance operator $\mathcal{C}_{XY} \colon \mathcal{H}^\prime \to \mathcal{H}$ with
\begin{equation}
\begin{split}
    \mathcal{C}_{XY}
    & \coloneqq \mathbb{E} \left[ \left( \phi \left( X \right) - \mathbb{E} \left[ \phi \left( X \right) \right] \right) \otimes \left( \phi^\prime \left( Y \right) - \mathbb{E} \left[ \phi^\prime \left( Y \right) \right] \right) \right].
\end{split}
\end{equation}
The Hilbert-Schmidt norm of an operator $\mathcal{C} \colon \mathcal{H}^\prime \to \mathcal{H}$ of Hilbert spaces $\mathcal{H}$ and $\mathcal{H}^\prime$ with orthonormal bases $a_1, a_2, \dots$ and $b_1, b_2, \dots$ is defined via $\left\lVert \mathcal{C} \right\rVert_{\mathrm{HS}} = \sum_{ij} \left\langle a_i, \mathcal{C} b_j \right\rangle_{\mathcal{H}}$ \citep{gretton2005measuring}.
It reduces to the Frobenius norm if both spaces are finite-dimensional Euclidean spaces.
For $g \in \mathcal{H}$ and $h \in \mathcal{H}^\prime$ it holds $\left\lVert g \otimes h \right\rVert_{\mathrm{HS}} = \left\lVert g \right\rVert_{\mathcal{H}} \left\lVert h \right\rVert_{\mathcal{H}^\prime}$, from which follows that 
\begin{equation}
\begin{split}
    \!\!\! \left\lVert \mathcal{C}_{XY} \right\rVert_{\mathrm{HS}}^2 & = \mathbb{E}_{X,Y,X^\mathrm{c},Y^\mathrm{c}} \left[ k \left( X, X^\mathrm{c} \right) k^\prime \left( Y, Y^\mathrm{c} \right) \right] - \mathbb{E}_{X,X^\mathrm{c},Y^\mathrm{c}} \left[ k \left( X, X^\mathrm{c} \right) \mathbb{E}_Y \left[ k^\prime \left( Y, Y^\mathrm{c} \right) \right] \right] \\
    & \quad - \mathbb{E}_{Y,X^\mathrm{c},Y^\mathrm{c}} \left[ \mathbb{E}_X \left[ k \left( X, X^\mathrm{c} \right) \right] k^\prime \left( Y, Y^\mathrm{c} \right) \right] + \mathbb{E}_{X,X^\mathrm{c}} \left[ k \left( X, X^\mathrm{c} \right) \right]\mathbb{E}_{Y,Y^\mathrm{c}} \left[ k^\prime \left( Y, Y^\mathrm{c} \right) \right],
\label{eq:HSIC_as_kernel}
\end{split}
\end{equation}
where $\left(X^\mathrm{c}, Y^\mathrm{c} \right)$ is an i.i.d. copy of $\left(X, Y \right)$ \citep{gretton2005measuring}.
Then, \cite{gretton2005measuring} show that it holds
\begin{equation}
\begin{split}
    & \left\lVert \mathcal{C}_{XY} \right\rVert_{\mathrm{HS}} = 0
    \iff \mathbb{E} \left[ \phi \left( X \right) \otimes \phi^\prime \left( Y \right) \right] = \mu_{\mathbb{P}_X} \otimes \mu_{\mathbb{P}_Y}^\prime %\mathbb{E} \left[ \phi \left( X \right) \right] \otimes \mathbb{E} \left[ \phi^\prime \left( Y \right) \right]
\end{split}
\end{equation}
with $\mu_{\mathbb{P}_Y}^\prime = \mathbb{E} \left[ \phi^\prime \left( Y \right) \right]$.
In consequence, they refer to $\operatorname{HSIC}_{k, k^\prime} \left( \mathbb{P}_{XY} \right) \coloneqq \left\lVert \mathcal{C}_{XY} \right\rVert_{\mathrm{HS}}^2$ as \textbf{Hilbert-Schmidt independence criterion} (HSIC).
Equation~\ref{eq:HSIC_as_kernel} indicates that we can estimate the HSIC in practice via the kernel trick, even when $\mathcal{H}$ or $\mathcal{H}^\prime$ are infinite-dimensional.
For two sets of samples $\mathbf{X} \coloneqq \left\{ X_1, \dots, X_n \right\}$ and $\mathbf{Y} \coloneqq \left\{ Y_1, \dots, Y_n \right\}$ with i.i.d. $\left(X_1, Y_1 \right), \dots, \left(X_n, Y_n \right) \sim \mathbb{P}_{XY}$ an estimator is given by
\begin{equation}
    \operatorname{HSIC}_{k, k^\prime} \left( \mathbf{X}, \mathbf{Y} \right) \coloneqq \operatorname{tr} \left( \mathbf{K}_X \left( I - \frac{1}{n}\mathbf{1}\mathbf{1}^\intercal \right) \mathbf{K}_Y \left( I - \frac{1}{n}\mathbf{1}\mathbf{1}^\intercal \right) \right),
\end{equation}
where $\left[ \mathbf{K}_X \right]_{ij} \coloneqq k \left( X_i, X_j \right)$, $\left[ \mathbf{K}_Y \right]_{ij} \coloneqq k^\prime \left( Y_i, Y_j \right)$, $\mathbf{1} = \left( 1, \dots, 1 \right)^\intercal$ is the vector of 1's, and $I$ is the unit matrix \citep{cortes2012algorithms}. \\
However, when we evaluate HSIC in practice, the estimated values will never be precisely zero.
This makes comparing HSIC values across different random variables problematic since their ranges may have different magnitudes.
%scales with the $X$ and $Y$ specific norms $\left\lVert \mathcal{C}_{XX} \right\rVert_{\mathrm{HS}}$ and $\left\lVert \mathcal{C}_{YY} \right\rVert_{\mathrm{HS}}$
Consequently, we use the normalized version of HSIC referred to as \textbf{central kernel alignment} (CKA) \citep{cortes2012algorithms, chang2013canonical}, which is defined by
\begin{equation}
\begin{split}
    \operatorname{CKA}_{k,k^\prime} \left( \mathbb{P}_{XY} \right) \coloneqq \frac{\left\lVert \mathcal{C}_{XY} \right\rVert_{\mathrm{HS}}^2}{\left\lVert \mathcal{C}_{XX} \right\rVert_{\mathrm{HS}}\left\lVert \mathcal{C}_{YY} \right\rVert_{\mathrm{HS}}}.
\end{split}
\end{equation}
It has the form of a squared correlation coefficient and by the Cauchy-Schwartz inequality lies within $\left[0, 1 \right]$ \citep{chang2013canonical}.
Even though $\operatorname{CKA}_{k,k^\prime}$ and $\operatorname{CMS}_k$ may appear similar in form, they measure very different things: While $\operatorname{CKA}_{k,k^\prime}$ measures the independence between random variables according to their joint distribution, $\operatorname{CMS}_k$ compares how similar their marginal distributions are via their mean embedding.
For two sets of samples $\mathbf{X} \coloneqq \left( X_1, \dots, X_n \right)$ and $\mathbf{Y} \coloneqq \left( Y_1, \dots, Y_n \right)$ with i.i.d. $\left(X_1, Y_1 \right), \dots, \left(X_n, Y_n \right) \sim \mathbb{P}_{XY}$ an estimator for $\operatorname{CKA}_{k,k^\prime} \left( \mathbb{P}_{XY} \right)$ is given by re-using the HSIC estimator via
\begin{equation}
    \widehat{\operatorname{CKA}}_{k,k^\prime} \left( \mathbf{X}, \mathbf{Y} \right) \coloneqq \frac{\operatorname{HSIC}_{k,k^\prime} \left( \mathbf{X}, \mathbf{Y} \right)}{\sqrt{\operatorname{HSIC}_{k,k^\prime} \left( \mathbf{X}, \mathbf{X} \right) \operatorname{HSIC}_{k,k^\prime} \left( \mathbf{Y}, \mathbf{Y} \right)}}.
\label{eq:CKA_est}
\end{equation}
Similar to the HSIC, it holds
\begin{equation}
    \operatorname{CKA}_{k,k^\prime} \left( \mathbb{P}_{XY} \right) = 0 \iff \mathbb{E}_{X,Y \sim \mathbb{P}_{XY}} \left[ \phi \left( X \right) \otimes \phi^\prime \left( Y \right) \right] = \mu_{\mathbb{P}_X} \otimes \mu_{\mathbb{P}_Y}^\prime. %\mathbb{E} \left[ \phi \left( X \right) \right] \otimes \mathbb{E} \left[ \phi^\prime \left( Y \right) \right] %
\label{eq:CKA0_implies_disentangl}
\end{equation}

In the next section, we state our theoretical main contribution, which uses the CKA to disentangle the CMS value of an entire target domain into the product of CMS values in sub-domains.

\section{Cosine Similarity of Disentangled Mean Embeddings}
\label{sec:cms_disent}

We can now state the main theoretical contribution of this work, which describes when we are allowed to disentangle the image-wise $\operatorname{CMS}$ into the product of more fine-grained cluster-wise $\operatorname{CMS}$ values.
\begin{theorem}
\label{th:cms_disentangl}
    Assume for random variables $X = \left( X_1, \dots, X_d \right)^\intercal$ and $Y = \left( Y_1, \dots, Y_d \right)^\intercal$ with outcomes in a space $\mathcal{X}^d$ and for a p.s.d. kernel $k \colon \mathcal{X} \times \mathcal{X} \to \mathbb{R}$, there exists a partition $\mathbf{I}$ of the indices $\left\{ 1 \dots d \right\}$ such that for all $I,I^\prime \in \mathbf{I}$ it holds $\operatorname{CKA}_{k^{\otimes \lvert I \rvert},k^{\otimes \lvert I^\prime \rvert}} \left( \mathbb{P}_{X_I X_{I^\prime}} \right) = 0 = \operatorname{CKA}_{k^{\otimes \lvert I \rvert},k^{\otimes \lvert I^\prime \rvert}} \left( \mathbb{P}_{Y_{I} Y_{I^\prime}} \right)$ with ${X_{I}} \coloneqq \left( X_i \right)_{i \in I}^\intercal$ and ${Y_{I^\prime}} \coloneqq \left( Y_i \right)_{i \in I^\prime}^\intercal$. Then, we have
    \begin{equation}
    \label{eq:cms_disentangl}
        \operatorname{CMS}_{k^{\otimes d}} \left( \mathbb{P}_X, \mathbb{P}_Y \right) = \prod_{I \in \mathbf{I}} \operatorname{CMS}_{k^{\otimes \lvert I \rvert}} \left( \mathbb{P}_{X_{I}}, \mathbb{P}_{Y_{I}} \right).
    \end{equation}
\end{theorem}
The proof located in Appendix~\ref{app:proves} is mostly based on Equation~\ref{eq:CKA0_implies_disentangl}.
\begin{algorithm}[tb]
   \caption{Monitoring Cosine Similarity of Disentangled Mean Embeddings}
   \label{alg:disent_cms}
\begin{algorithmic}
    \STATE {\bfseries Input:} Training data $D^{\mathrm{tr}} \in \mathbb{R}^{n_{\mathrm{tr}} \times wh}$ and test data $D^{\mathrm{te}} \in \mathbb{R}^{n_{\mathrm{te}} \times wh}$ with $n_{\mathrm{tr}}$ and $n_{\mathrm{te}}$ number of flattened pixels with resolution $w \times h$, generator $G_t$ with training iterations $t \in \mathbb{N}$ and $n^\prime$ image generations, p.s.d. kernel $k \colon \mathbb{R} \times \mathbb{R} \to \mathbb{R}$.
    \STATE Initialize empty correlation matrix $M \in \mathbb{R}^{wh \times wh}$.
    \FOR{$i = 1, \dots, wh$}
        \FOR{$j = 1, \dots, wh$}
         \STATE \COMMENT{ Compute centered kernel alignment between pixel $i$ and $j$}
         \STATE $M_{ij} \gets \widehat{\operatorname{CKA}}_{k,k} \left( \left\{ D_{li}^{\mathrm{tr}} \right\}_{l = 1 \dots n_{\mathrm{tr}}}, \left\{ D_{lj}^{\mathrm{tr}} \right\}_{l = 1 \dots n_{\mathrm{tr}}} \right)$ according to Eq.~\ref{eq:CKA_est}
        \ENDFOR
    \ENDFOR
   \STATE $\left\{I_1, \dots, I_C \right\} \leftarrow \operatorname{HierarchicalClustering} \left( M \right)$ 
   \FOR{$t=1, \dots $}
   \STATE $G_{t+1} \gets \operatorname{TrainingIteration} \left( G_t \right)$
   \STATE $\hat{D} \leftarrow$ generate $n^\prime$ images with $G_{t+1}$
   \STATE $\text{ImageSimilarity} \gets \widehat{\operatorname{CMS}}_{k^{\otimes wh}} \left( \left\{ D_{ij}^{\mathrm{te}} \right\}_{i=1 \dots n_{\mathrm{te}}, j=1 \dots wh}, \{ \hat{D}_{ij} \}_{i=1 \dots n^\prime, j=1 \dots wh} \right)$
   \FOR{$I, c$ {\bfseries in} $\operatorname{enumerate} \left\{I_1, \dots, I_C \right\}$}
   \STATE $\text{ClusterSimilarity}_c \gets \widehat{\operatorname{CMS}}_{k^{\otimes \lvert I \rvert}} \left( \left\{ D_{ij}^{\mathrm{te}} \right\}_{i=1\dots n_{\mathrm{te}}, j \in I}, \{ \hat{D}_{ij} \}_{i=1\dots n^\prime, j \in I} \right)$
   \ENDFOR
   \STATE Output $\text{ImageSimilarity}, \left\{ \text{ClusterSimilarity}_c \right\}_{c=1\dots C}$
   \ENDFOR
\end{algorithmic}
\end{algorithm}
% while the same does not hold for the MMD since
% \begin{equation}
% \begin{split}
%     \operatorname{MMD}_k^2 \left(P, Q \right) & = \left\lVert \bigotimes_{c=1}^C \mu_{I_c} \left( P \right) - \bigotimes_{c=1}^C \mu_{I_c} \left( Q \right) \right\rVert_{\mathcal{H}_{\operatorname{img}}}^2 \neq \underbrace{\prod_{c=1}^C \left\lVert \mu_{I_c} \left( P \right) - \mu_{I_c} \left( Q \right) \right\rVert_{\mathcal{H}_{\operatorname{img}}}^2}_{ = \prod_{c=1}^C \operatorname{MMD}^2 \left(P_{I_c}, Q_{I_c} \right)}. \\
% \end{split}
% \end{equation}
% However, if there exists $i \in \left\{1 \dots d \right\}$ such that $\operatorname{MMD}_{k_{\operatorname{pxl}}}^2 \left(P_i, Q_i \right) = 0$ then if $C \to d$ it holds
% \begin{equation}
%     \prod_{c=1}^C \operatorname{MMD}_{k_{I_c}}^2 \left(P, Q \right) \rightarrow 0.
% \end{equation}
% Consequently, disentangling the MMD is not a good approach to approximate the image-wise MMD.
% However tracking the individual cluster-wise MMD values 
Theorem~\ref{th:cms_disentangl} indicates, that, by finding appropriate clusters, we can track the individual cluster-wise $\operatorname{CMS}$ values without losing information about the overall image-wise $\operatorname{CMS}$.
The cluster-wise $\operatorname{CMS}$ values then help to identify the cluster(s) responsible for certain behavior of the overall image-wise $\operatorname{CMS}$, improving the interpretability of the model performance and the training dynamics.
One practical constraint is that the estimated CKA values are likely not exactly zero since estimators cannot be expected to be perfectly precise and there may often be an infinitesimal correlation between pixels.
This indicates that the assumptions of Theorem~\ref{th:cms_disentangl} will be, strictly speaking, violated to some degree in practice.
However, it is straightforward to verify if the disentanglement is meaningful by comparing both sides of Equation~\ref{eq:cms_disentangl}, as we will see in Section~\ref{sec:exp}.

If we want to turn Theorem~\ref{th:cms_disentangl} into a practical algorithm, we are facing a runtime problem:
The number of possible partitions for a set grows exponentially with the set size \citep{berend2010improved}, which makes evaluating the $\operatorname{CKA}$ for all possible partitions of an image grid infeasible.
As a workaround, we only compute the $\operatorname{CKA}$ between all pairwise pixels, which has $O \left( d^2 \right)$ runtime complexity.
We then perform hierarchical clustering to identify clusters of pixels with high pairwise $\operatorname{CKA}$ values.
Further, we only compute the CKA values based on the training data as the generated images converge to this distribution during training.
As we will see in the experiments, this simple approach works sufficiently well to find meaningful clusters.
The whole algorithm to disentangle the $\operatorname{CMS}$ for monitoring the training performance of an image generator is presented in Algorithm~\ref{alg:disent_cms}, where we use the CMS estimator of Equation~\ref{eq:CMS_est} and the CKA estimator of Equation~\ref{eq:CKA_est}.

\begin{figure*}[t]
\centering
\includegraphics[width=0.8\columnwidth]{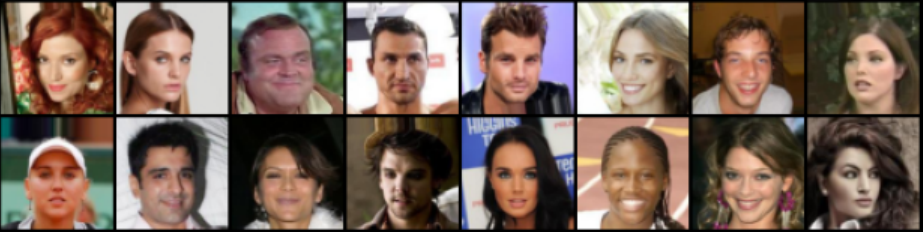}
\caption[]{
    Samples of the CelebA dataset. Most faces are centered of similar size and similar angles. The clusters identified in Figure~\ref{fig:clustering} match this observation.
}
\label{fig:celeba_samples}
\end{figure*}

\begin{figure*}[t]
\centering
    \begin{subfigure}{.51\textwidth}
    \centering
    \includegraphics[width=\columnwidth]{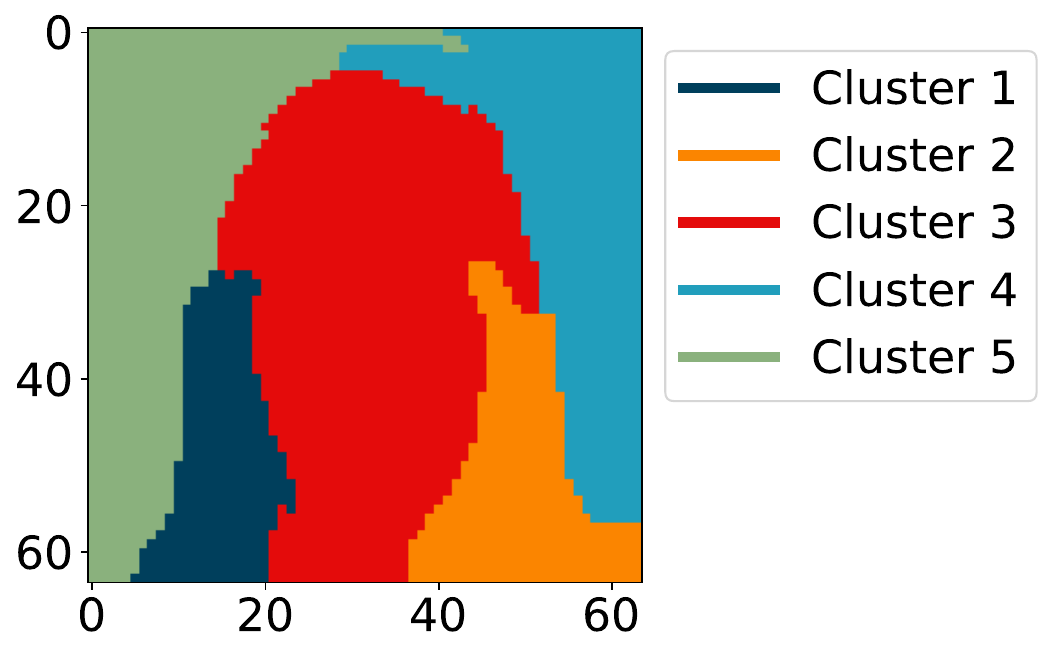}
    \caption{Identified clusters in pixel grid}
    \end{subfigure}%
    \begin{subfigure}{.41\textwidth}
    \centering
    \includegraphics[width=\columnwidth]{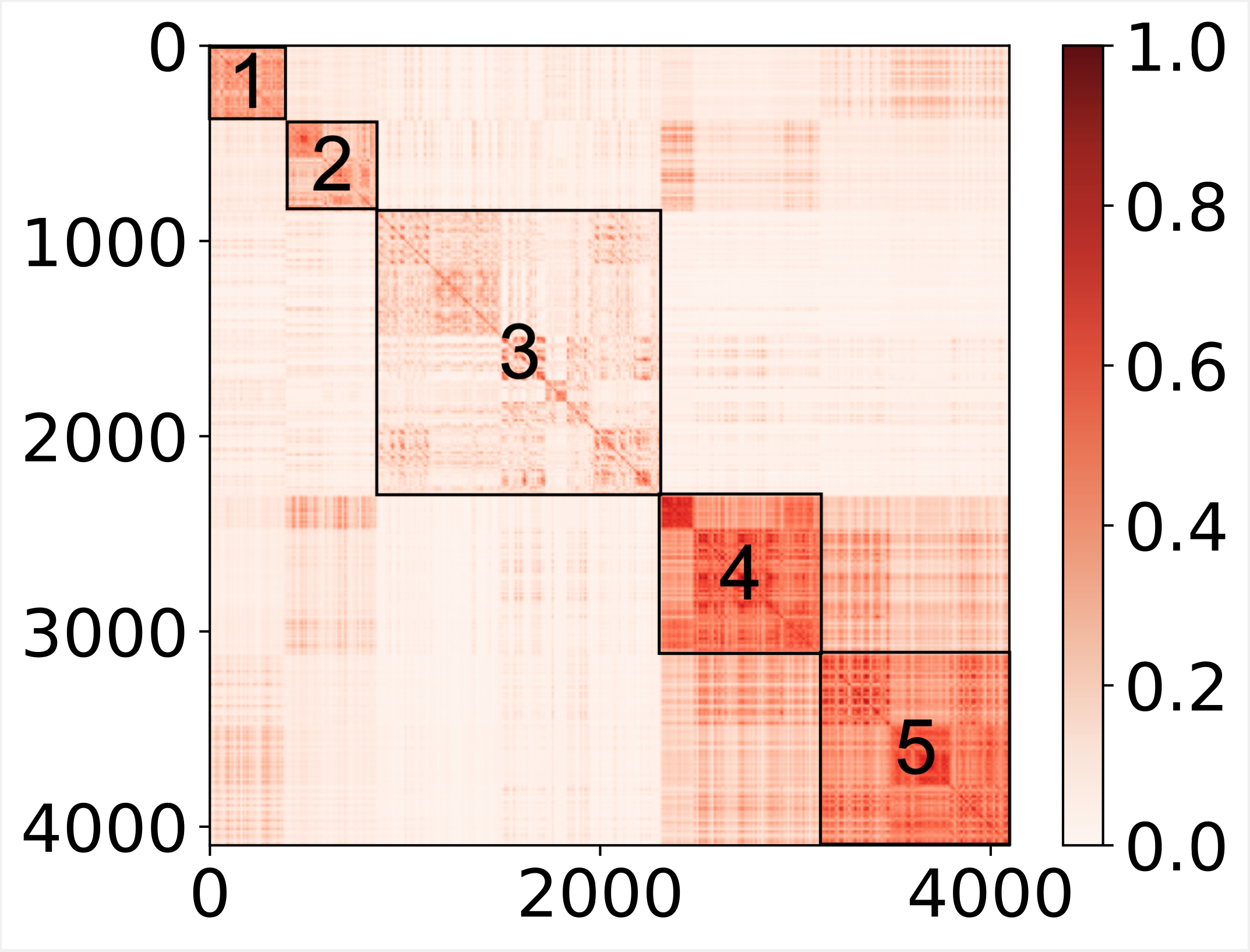}
    \caption{$\operatorname{CKA}$ matrix of flattened pixels}
    \end{subfigure} \\
    \begin{subfigure}{.78\textwidth}
    \centering
    \includegraphics[width=\columnwidth]{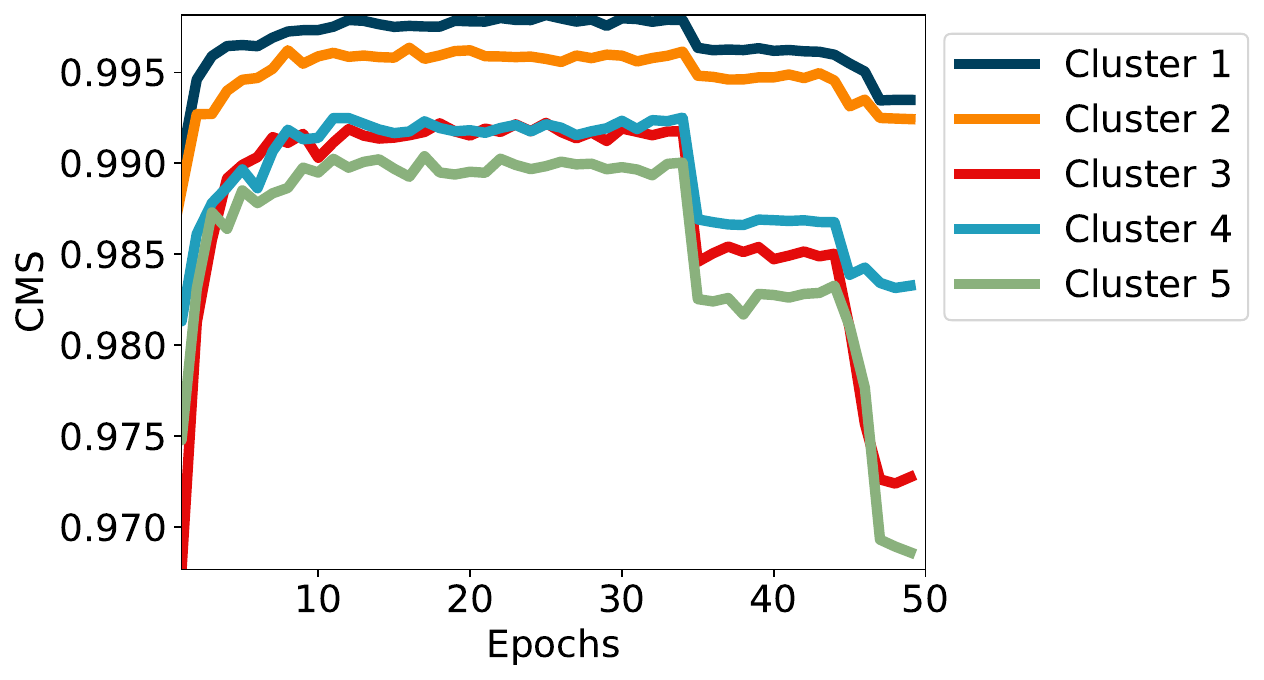}
    \caption{$\operatorname{CMS}$ values throughout training}
    \label{fig:cluster_wise_CMS}
    \end{subfigure} \\

\caption[]{
    \textbf{Top-Left:} The identified clusters match how a human may separate the image structure of CelebA: There are two clusters for the background (Clusters 4 \& 5), two clusters for long hair or alternating head angles (Clusters 1 \& 2), and one central cluster for the head and neck (Cluster 3).
    \textbf{Top-Right:} The correlation matrix in terms of the $\operatorname{CKA}$ values indicates how well the clusters can be separated. The blocks on the diagonal are ordered by cluster number. As can be seen, most clusters are fairly independent of the other clusters (especially Cluster 3). Only clusters 4 and 5 show a relatively strong dependence on each other, which is expected since these often express the same background in the images (c.f. Figure~\ref{fig:celeba_samples}).
    \textbf{Bottom:} Unlike the other errors, we can decompose the image-wise $\operatorname{CMS}$ into the $\operatorname{CMS}$ of different clusters according to the $\operatorname{CKA}$. This offers novel insights into model performance.
    For example, we can detect that Cluster 3 and 5 degrade more in the training collapses than the other clusters.
}
\label{fig:clustering}
\end{figure*}

\begin{figure*}[t]
\centering
    \begin{subfigure}{.47\textwidth}
    \centering
    \includegraphics[width=\columnwidth]{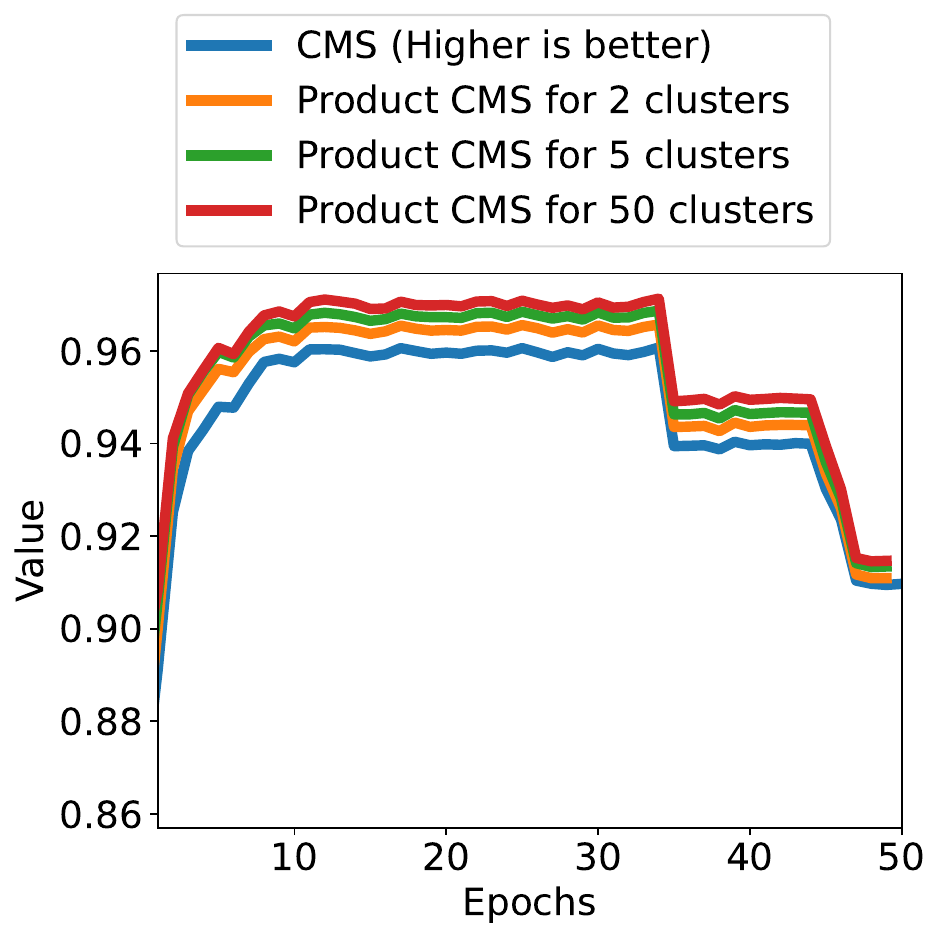}
    \end{subfigure}%
    \begin{subfigure}{.53\textwidth}
    \centering
    \includegraphics[width=\columnwidth]{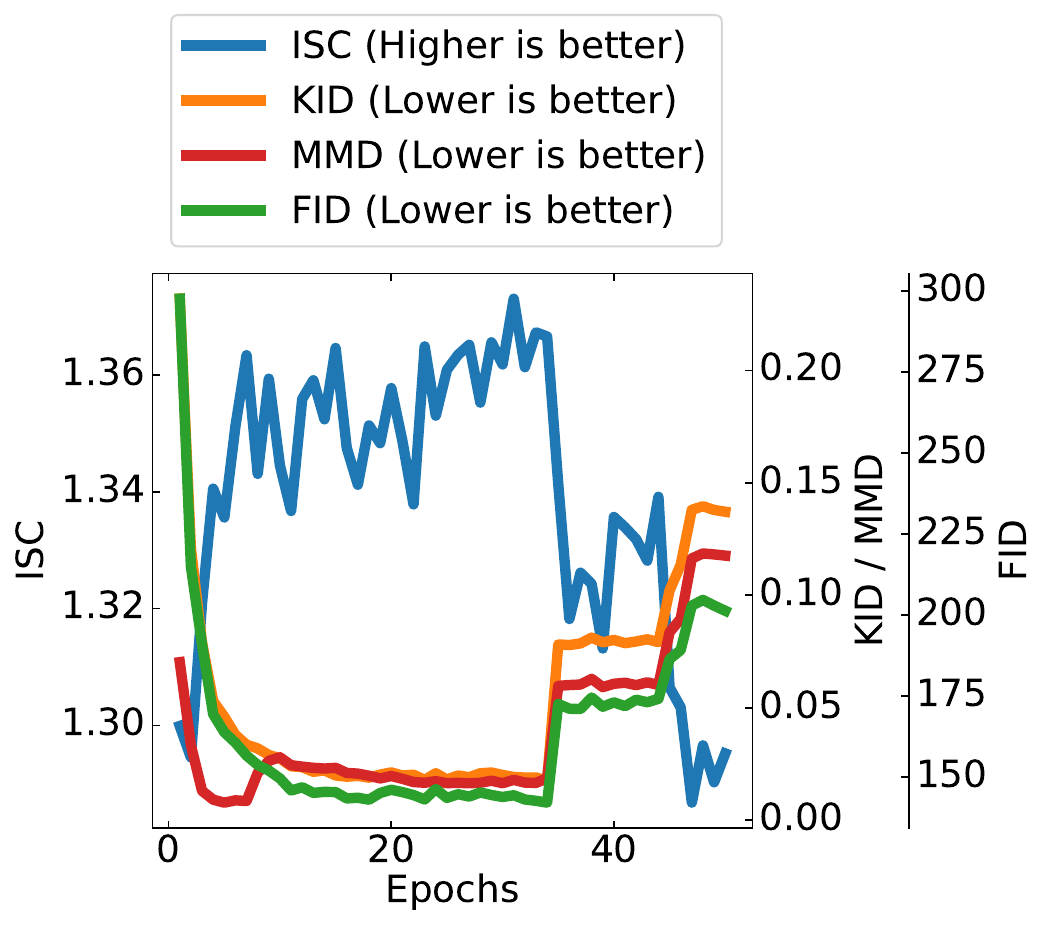}
    \end{subfigure}
\caption[]{
    Different errors throughout the training of a DCGAN model on the CelebA dataset. All lines are an average of 20 seeds. \textbf{Left:} The $\operatorname{CMS}$ (higher is better) shows how the average training run improves until epoch 10. After epoch 30, some models collapse, and after epoch 45 additional collapses occur. Computing the product of the cluster-wise $\operatorname{CMS}$ values according to our methodology shows a close match with the normal $\operatorname{CMS}$, indicating the correctness of the clusters.
    \textbf{Right:} The ISC does match the other errors but shows erratic behavior.
    The KID and FID resemble the $\operatorname{CMS}$ quite closely.
    The $\operatorname{MMD}$ also shows a similar trajectory as the other errors but indicates a minimum around epochs 5-7.
    Generated samples of the training runs match these observations (c.f. Figure~\ref{fig:gen_samples}).
}
\label{fig:image_wise_errors}
\end{figure*}

% \begin{figure*}[t]
% \centering
% \includegraphics[width=0.8\columnwidth]{Styles/Figures/Cluster cos of DCGAN64x64 0 batch_size=128 ngf=64 ndf=64 lr=0.0002 beta1=0.5 rbf_kernel.pdf}
% \caption[]{
%     Unlike the other errors, we can decompose the image-wise $\operatorname{CMS}$ into the $\operatorname{CMS}$ of different clusters according to the $\operatorname{CKA}$. This offers novel insights into the source of errors. For example, we can detect that Cluster 4 (the face area colored in red) degrades more in the training collapses as the other cluster.
% }
% \label{fig:cluster_wise_CMS}
% \end{figure*}

\section{Experiments}
\label{sec:exp}

The source code for the following experiments is located at \url{https://github.com/MLO-lab/Disentangling_Mean_Embeddings}.
We run experiments on the CelebA dataset \citep{liu2015deep}, which consists of 200.000 colored celebrity images with resolution 64 x 64, and on the ChestMNIST dataset \citep{medmnistv1} consisting of 112.120 gray-scale chest scans with 28 x 28 resolution. Since both datasets show centered faces/chests at a similar angle, we can expect to identify various clusters of pixels that can be interpreted in a meaningful way (c.f. Figure~\ref{fig:celeba_samples} and Figure~\ref{fig:gen_samples_chest_dcgan} for samples).
For CelebA, we train 20 seeds of the DCGAN architecture \citep{radford2015unsupervised} on randomly sampled 90\% of the original set, and use the other 10\% for evaluation.
As errors, we consider the $\operatorname{CMS}$ and $\operatorname{MMD}$ based on the RBF kernel with $\gamma$ set to the inverse of the median of the pairwise Euclidean distances between training instances, which is a heuristic based on \citep{scholkopf2002learning}.
Further, we evaluate the Inception Score Criterion (ISC) \citep{salimans2016improved}, Fréchet Inception Distance (FID) \citep{heusel2017gans}, and the Kernel Inception Distance (KID) \citep{binkowski2018demystifying}.
We average all errors across all seeds. More details are given in Appendix~\ref{app:exp}. \\
In Figure~\ref{fig:clustering}, we show the identified clusters for CelebA on the left, and the respective pairwise $\operatorname{CKA}$ values of the (flattened) pixels on the right, which we refer to as $\operatorname{CKA}$ matrix.
The indices in the $\operatorname{CKA}$ matrix are arranged according to the clusters.
As can be seen, Cluster 3, which represents the face, is fairly independent of the other clusters.
However, Cluster 4 and 5 share a lot of dependence, which is not surprising, since these represent the background.
The training curves for the DCGAN architecture are depicted in Figure~\ref{fig:image_wise_errors}.
There, on the left, we compare the image-wise $\operatorname{CMS}$ with the product of the cluster-wise $\operatorname{CMS}$ to verify that Theorem~\ref{th:cms_disentangl} holds approximately.
On the right in Figure~\ref{fig:image_wise_errors}, we show the ISC, FID, KID, and $\operatorname{MMD}$ for comparison.
All metrics capture similar trends in most cases.
The benefits of our approach become apparent in Figure~\ref{fig:cluster_wise_CMS}, where we plot the cluster-wise $\operatorname{CMS}$ values for the detected clusters.
Here, we can determine how much each cluster influences the image-wise $\operatorname{CMS}$ throughout training.
Especially Cluster 3 and 5, which represent the head and left background area, are striking: They degrade the worst among all clusters after the two training collapses.
This indicates, that these pixel regions have the highest influence on the worsening image-wise $\operatorname{CMS}$. \\
In Appendix~\ref{app:exp}, we discuss the results for ChestMNIST in more detail.
Specifically, we compare the DCGAN with the DDPM architecture \citep{ho2020denoising} in Figure~\ref{fig:clustering_Chest}.
There, we discover how a major performance drop during the training runs with the DCGAN architecture can be assigned to the background of the images.
On the contrary, the DDPM architecture fits all regions quickly except the background region, which requires further iterations. \\
Overall, such an analysis is only possible with our approach and the $\operatorname{CMS}$ as error, since the other errors ($\operatorname{MMD}$, FID, KID, ISC) cannot be disentangled similarly.

\paragraph{Limitations.} Our approach offers novel insights, but it assumes mean embeddings are a meaningful representation of the images based on a user-defined product kernel. While kernels scale well to higher dimensions, they will still degrade at some resolution \citep{JMLR:v13:gretton12a}. Further, computing the clusters is computationally expensive and we may not expect to find perfectly independent clusters in practice. %Image generation evaluation is a difficult problem to which the $\operatorname{CMS}$ 

\section{Conclusion}

In this work, we introduced a novel approach to disentangle the mean embedding of an image space into mean embeddings of approximately independent pixel clusters.
We also proved when the cosine similarity of the mean embeddings can be disentangled into the product of the cosine similarities for each respective cluster.
This enables the evaluation and interpretation of the generalization performance of each cluster in isolation, significantly enhancing the explainability and the likelihood of identifying model misbehavior.
We demonstrated the improved interpretability by monitoring the training of various architectures on the CelebA and ChestMNIST datasets according to the MMD, ISC, FID, KID, and our approach.

\newpage

\bibliography{main}
\bibliographystyle{plainnat}

%%%%%%%%%%%%%%%%%%%%%%%%%%%%%%%%%%%%%%%%%%%%%%%%%%%%%%%%%%%%

\appendix

\newpage

\begin{figure*}[!ht]
\centering
    \begin{subfigure}{.5\textwidth}
    \centering
    \includegraphics[width=.9\columnwidth]{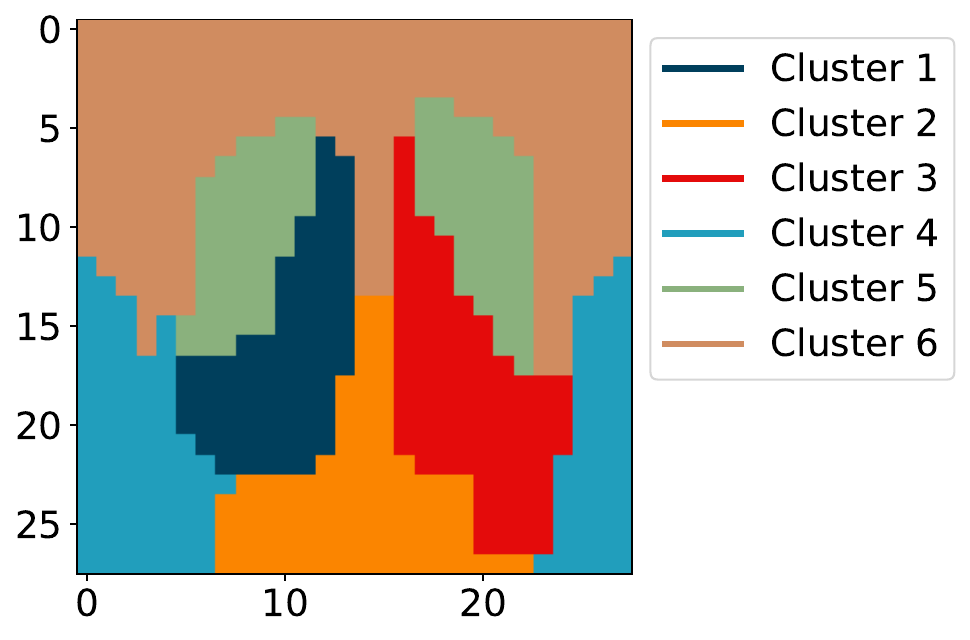}
    \caption{Identified clusters in pixel grid}
    \end{subfigure}%
    \begin{subfigure}{.45\textwidth}
    \centering
    \includegraphics[width=\columnwidth]{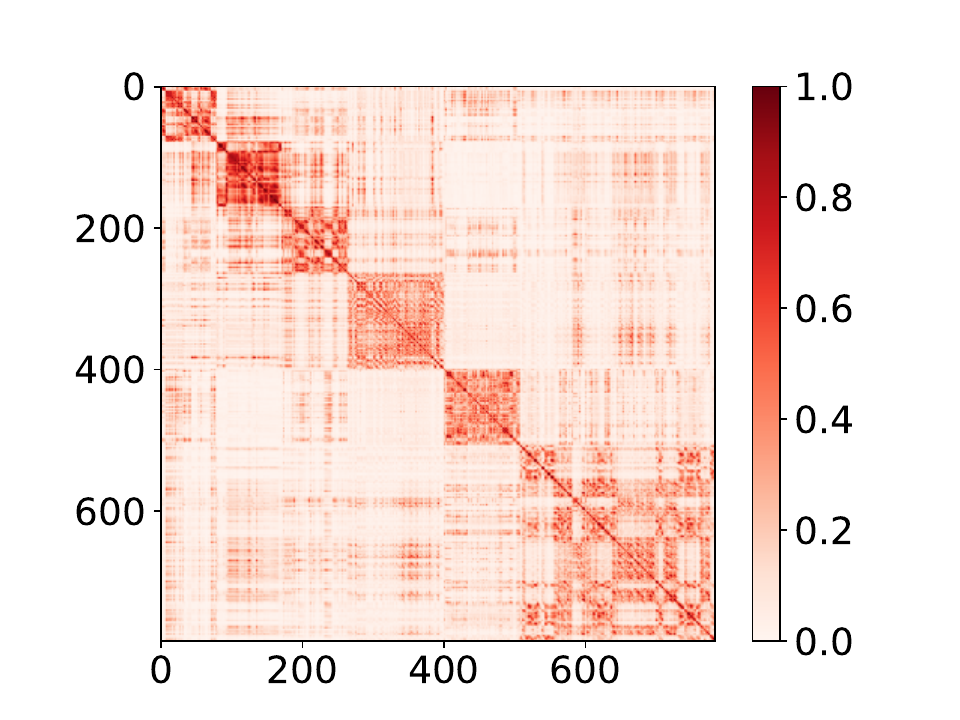}
    \caption{$\operatorname{CKA}$ matrix of flattened pixels}
    \end{subfigure} \\
    \begin{subfigure}{.5\textwidth}
    \centering
    \includegraphics[width=\columnwidth]{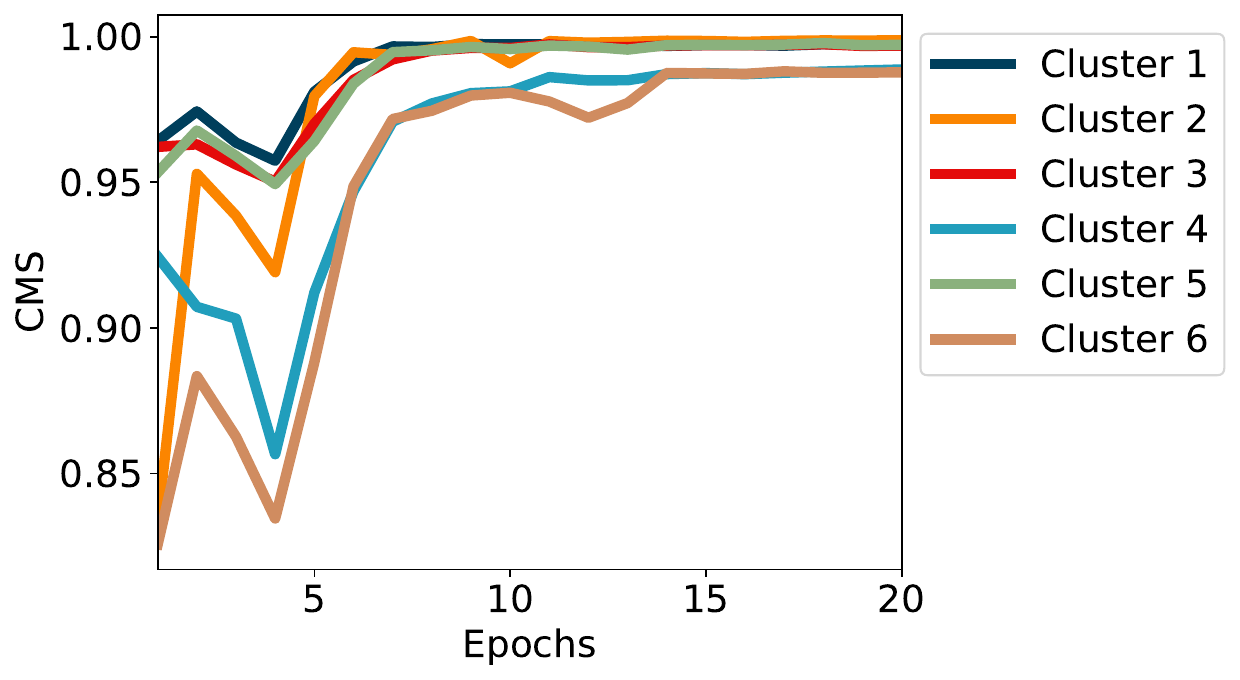}
    \caption{$\operatorname{CMS}$ values throughout training (DCGAN)}
    \label{fig:cluster_wise_CMS_Chest_DCGAN}
    \end{subfigure}%
    \begin{subfigure}{.5\textwidth}
    \centering
    \includegraphics[width=\columnwidth]{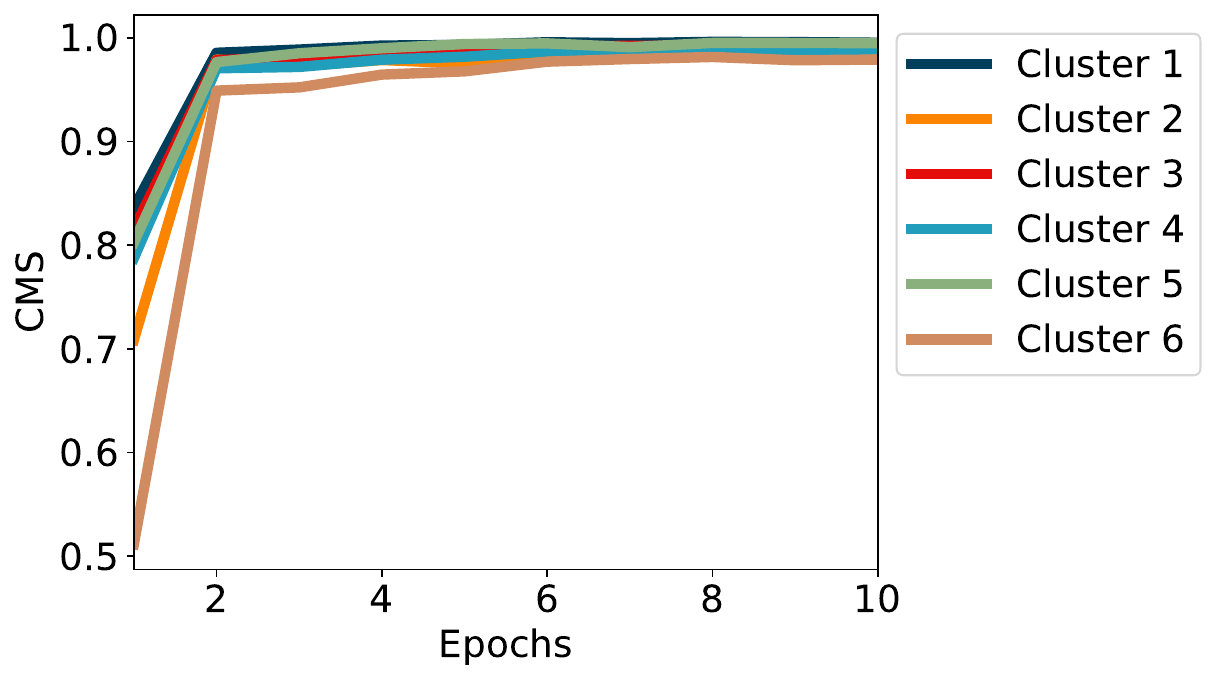}
    \caption{$\operatorname{CMS}$ values throughout training (DDPM)}
    \label{fig:cluster_wise_CMS_Chest_DDPM}
    \end{subfigure} \\
    \begin{subfigure}{.5\textwidth}
    \centering
    \includegraphics[width=.8\columnwidth]{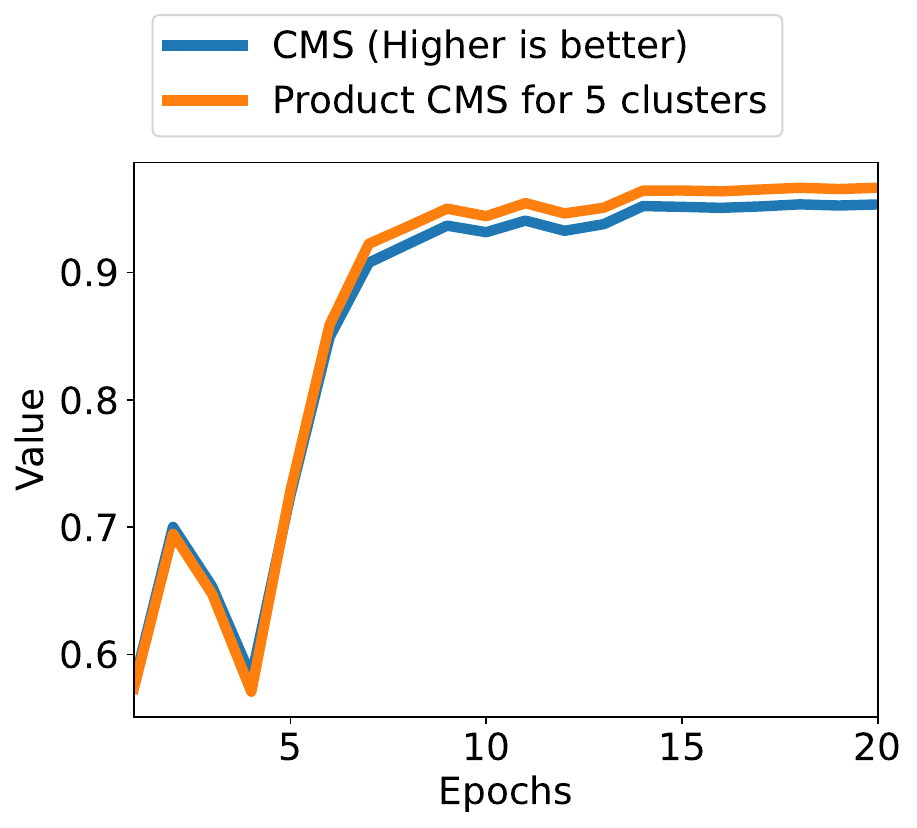}
    \caption{$\operatorname{CMS}$ values throughout training (DCGAN)}
    \label{fig:image_wise_CMS_Chest_DCGAN}
    \end{subfigure}%
    \begin{subfigure}{.5\textwidth}
    \centering
    \includegraphics[width=.8\columnwidth]{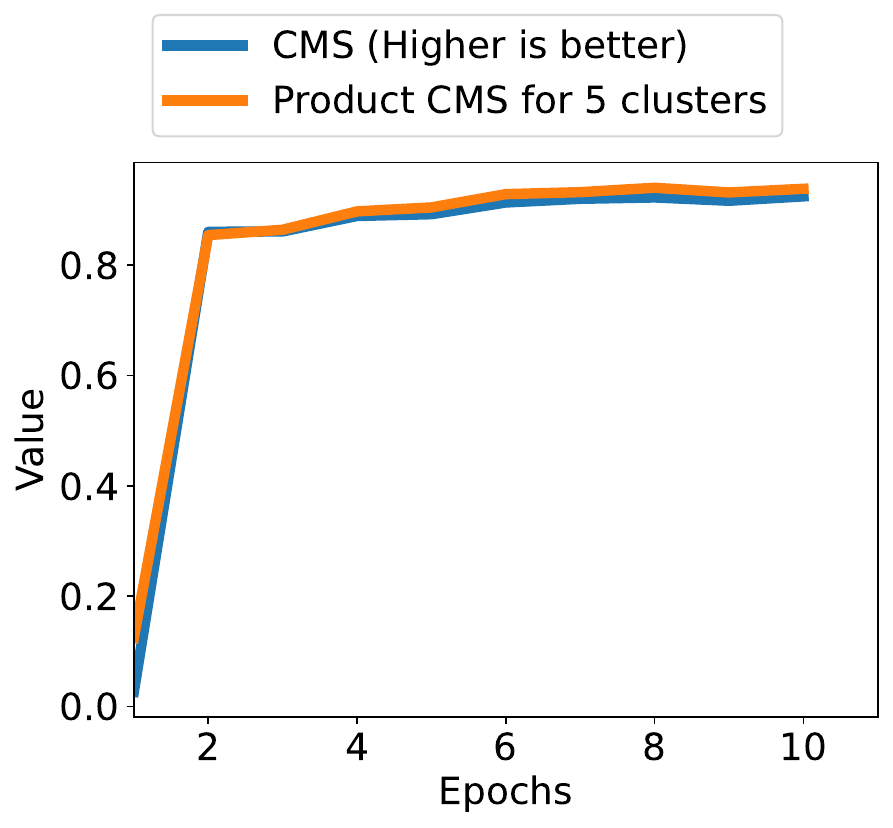}
    \caption{$\operatorname{CMS}$ values throughout training (DDPM)}
    \label{fig:image_wise_CMS_Chest_DDPM}
    \end{subfigure} \\
\caption[]{
    \textbf{Top-Left:} The identified clusters for ChestMNIST match how a human may separate the image structure: There are three clusters for the lung area, one for the abdomen, and two for the upper chest and background.
    \textbf{Top-Right:} The correlation matrix in terms of the $\operatorname{CKA}$ values indicates how well the clusters can be separated. The blocks on the diagonal are ordered by cluster number. As can be seen, most clusters are fairly independent. Cluster 6 could be further separated.
    \textbf{Mid:} Comparing the cluster-wise $\operatorname{CMS}$ values throughout training of DCGAN and DDPM architectures shows the difficulty of learning each cluster. The DCGAN architectures have a performance drop mostly due to Cluster 5 and 6 around Epoch 4.
    \textbf{Lower:} The cluster-wise $\operatorname{CMS}$ values successfully represent the image-wise $\operatorname{CMS}$.
}
\label{fig:clustering_Chest}
\end{figure*}

\begin{figure*}[t]
\centering
    \begin{subfigure}{.5\textwidth}
    \centering
    \includegraphics[width=\columnwidth]{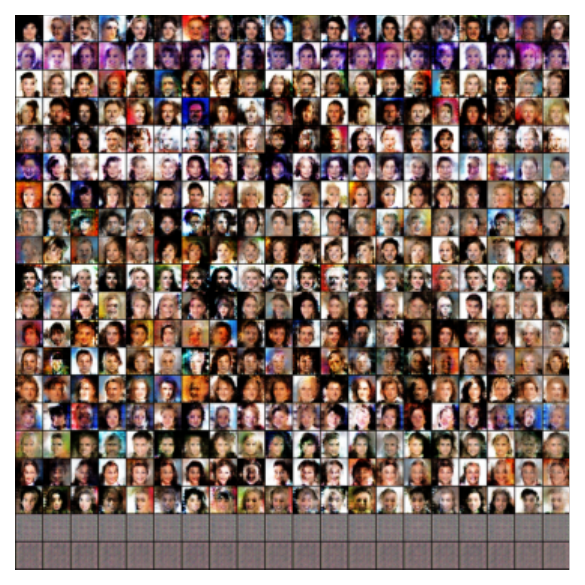}
    \caption{Epoch 1}
    \end{subfigure}%
    \begin{subfigure}{.5\textwidth}
    \centering
    \includegraphics[width=\columnwidth]{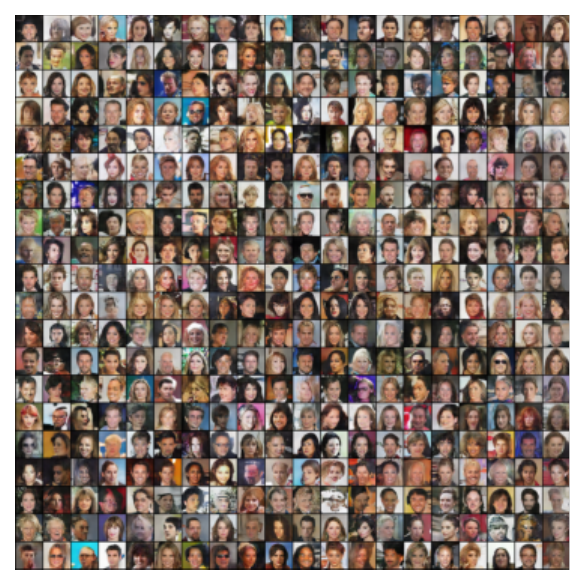}
    \caption{Epoch 21}
    \end{subfigure} \\
    \begin{subfigure}{.5\textwidth}
    \centering
    \includegraphics[width=\columnwidth]{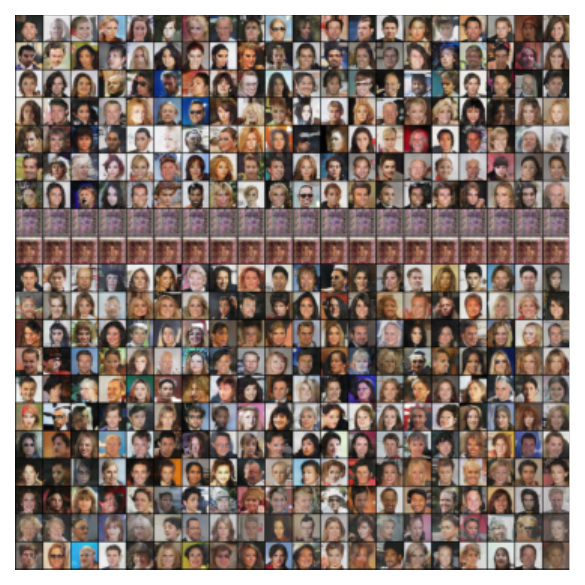}
    \caption{Epoch 41}
    \end{subfigure}%
    \begin{subfigure}{.5\textwidth}
    \centering
    \includegraphics[width=\columnwidth]{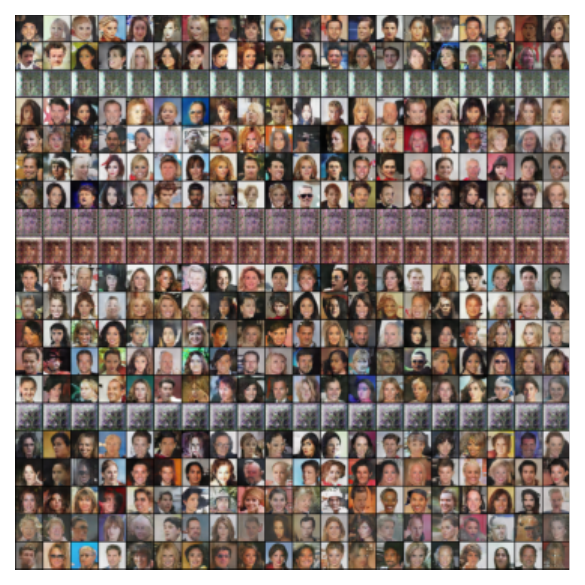}
    \caption{Epoch 49}
    \end{subfigure} \\
\caption[]{
    Generated samples of the twenty training runs (each row is a seed, each column a sample of a respective seed).
    Initially, all models are improving their fit.
    At 21 epochs, no further improvements are visible.
    At 41 epochs, the training of two models collapsed.
    At 49 epochs, the training of two additional models collapsed.
    The collapses are visible in all evaluation metrics in Figure~\ref{fig:image_wise_errors}, but only with our approach we can quantify the extend to which the individual pixel regions are affected.
}
\label{fig:gen_samples}
\end{figure*}

\begin{figure*}[t]
\centering
    \begin{subfigure}{.5\textwidth}
    \centering
    \includegraphics[width=\columnwidth]{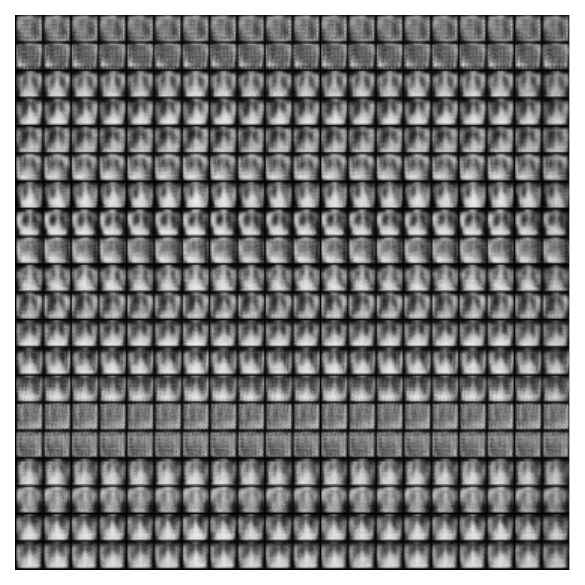}
    \caption{Epoch 3}
    \end{subfigure}%
    \begin{subfigure}{.5\textwidth}
    \centering
    \includegraphics[width=\columnwidth]{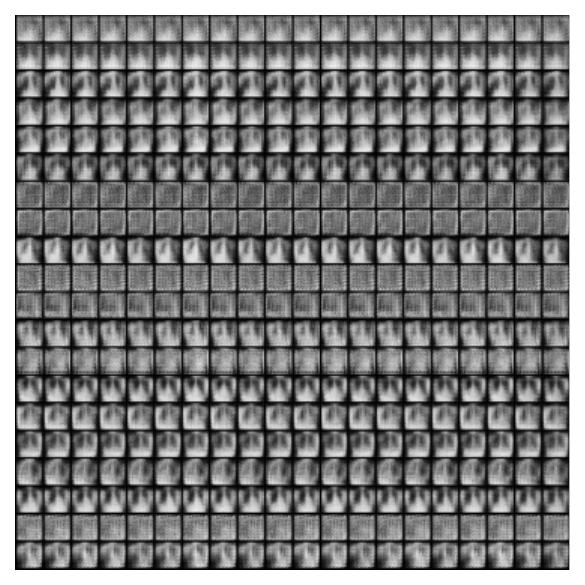}
    \caption{Epoch 4}
    \end{subfigure} \\
    \begin{subfigure}{.5\textwidth}
    \centering
    \includegraphics[width=\columnwidth]{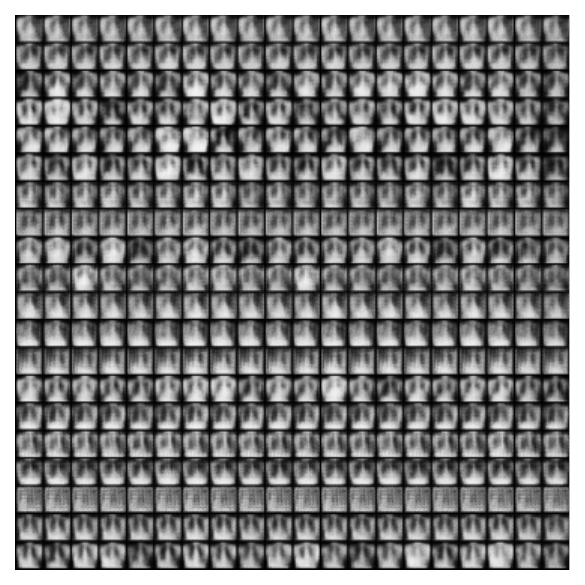}
    \caption{Epoch 5}
    \end{subfigure}%
    \begin{subfigure}{.5\textwidth}
    \centering
    \includegraphics[width=\columnwidth]{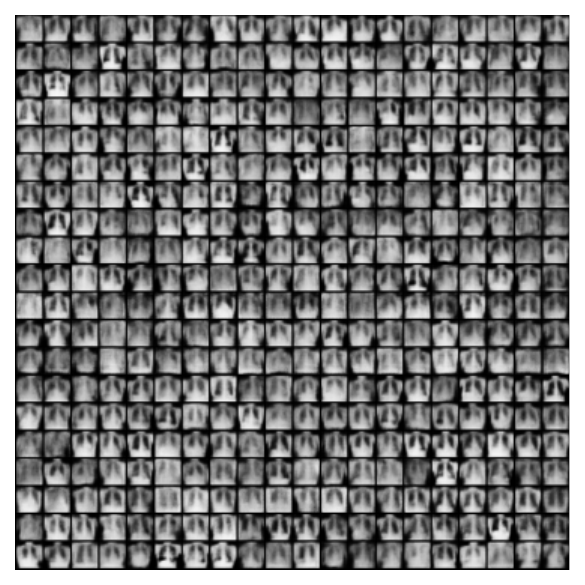}
    \caption{Epoch 10}
    \end{subfigure}\\
\caption[]{
    Generated samples of the twenty training runs (each row is a seed, each column a sample of a respective seed) of the DCGAN architecture on ChestMNIST.
}
\label{fig:gen_samples_chest_dcgan}
\end{figure*}

\section{Overview}
\label{app:overview}

In the following, we discuss additional experiment results and details in Appendix~\ref{app:exp}, practical time and space complexity of Algorithm~\ref{alg:disent_cms} in Appendix~\ref{app:time_space_compl}, and missing proofs in Appendix~\ref{app:proves}.

\section{Additional Experimental Results and Details}
\label{app:exp}

In this section, we give more experimental results and details, which are missing in the main part.

\subsection{ChestMNIST and additional CelebA Figures}

We train 20 seeds of the DCGAN and DDPM architecture on the provided training set of ChestMNIST.
Generated samples of the DCGAN architecture are presented in Figure~\ref{fig:gen_samples_chest_dcgan}.
In Figure~\ref{fig:clustering_Chest}, we show the corresponding plots of ChestMNIST as in Figure~\ref{fig:clustering}.
Specifically, we also discover a meaningful cluster partition of the image grid: It becomes clear in what regions the lungs, the abdomen, and the background are located.
The $\operatorname{CKA}$ matrix shows that the large background region in brown is sparse and may be split up in additional clusters.
The cluster-wise $\operatorname{CMS}$ values show how each architecture behaves during training: The DCGAN models suffer from a performance drop after 4 epochs, which can be assigned to Cluster 4 and 6 (the background).
The DDPM models are more stable but learning Cluster 6 takes longer than the other clusters.
The image-wise $\operatorname{CMS}$ values at the bottom of Figure~\ref{fig:clustering_Chest} indicate that the cluster-wise $\operatorname{CMS}$ values are representative for the image-wise $\operatorname{CMS}$.

% In Figure~\ref{fig:celeba_samples}, we show samples of the CelebaA dataset.
% It is visually clear that the identified clusters in Figure~\ref{fig:clustering} match meaningful pixel regions.
We also show samples of generated images in Figure~\ref{fig:gen_samples}, which matches the observed trends of the evaluation metrics in Figure~\ref{fig:image_wise_errors}.

\subsection{Experimental Details}

%The source code for all the conducted experiments is publicly available at \url{https://github.com/waiting-for-acceptance}.

All models were trained on a machine equipped with an AMD Ryzen 9 3950X CPU, an Nvidia RTX 4090 GPU, and 128GB of RAM.
However, it is important to note that such high-end hardware is not strictly necessary for training these models; similar results can be obtained on less powerful systems, albeit with potentially longer training times.
We use a random split of 90\% of the CelebA dataset for training, utilizing the original model architecture as described in \citep{radford2015unsupervised}.
The specific split can be reproduced via our source code.
Each model and training run is initialized with a unique random seed ranging from 0 to 20.

The models for CelebA were trained with a consistent set of hyperparameters: a batch size of 128, generator and discriminator feature maps set to 64, a learning rate of 0.0002, and Adam optimizer $\beta$ values of (0.5, 0.999). Binary Cross-Entropy (BCE) loss was used for training both the generator and discriminator. These settings were uniformly applied across all models.

For ChestMNIST we also train with a consistent set of hyperparameters: For the DCGAN architecture a batch size of 128, generator and discriminator feature maps set to 64, a learning rate of 1e-05, and Adam optimizer $\beta$ values of (0.5, 0.999). Binary Cross-Entropy (BCE) loss was used for training both the generator and discriminator. These settings were uniformly applied across all models. \\
For the DDPM architecture, we base our implementation on \url{https://github.com/tcapelle/Diffusion-Models-pytorch} with similar hyperparameters.

Similar to the $\operatorname{CKA}$, we also compute the $\operatorname{MMD}$ and $\operatorname{CMS}$ via their kernel representations according to the following.
Given two datasets $\mathbf{X} = \left( X_1, \dots X_n \right) \overset{iid}{\sim} \mathbb{P}_X$ and $\mathbf{Y} = \left( Y_1, \dots Y_m \right) \overset{iid}{\sim} \mathbb{P}_Y$, we use $\widehat{\lVert \mu_{\mathbb{P}_X} \rVert_{\mathcal{H}}^2} \coloneqq \frac{1}{n^2} \sum_{i,j=1}^n k \left( X_i, X_j \right)$ as estimator for $\lVert \mu_{\mathbb{P}_X} \rVert_{\mathcal{H}}^2$, $\widehat{\lVert \mu_{\mathbb{P}_Y} \rVert_{\mathcal{H}}^2} \coloneqq \frac{1}{m^2} \sum_{i,j=1}^m k \left( X_i, X_j \right)$ as estimator for $\lVert \mu_{\mathbb{P}_Y} \rVert_{\mathcal{H}}^2$, and $\widehat{\langle \mu_{\mathbb{P}_X}, \mu_{\mathbb{P}_Y} \rangle_{\mathcal{H}}} \coloneqq \frac{1}{nm} \sum_{i=1}^n \sum_{j=1}^m k \left( X_i, Y_j \right)$ as estimator for $\langle \mu_{\mathbb{P}_X}, \mu_{\mathbb{P}_Y} \rangle_{\mathcal{H}}$. \\
Based on \citep{JMLR:v13:gretton12a} and \citep{kubler2019quantum} we use these as plugins for the $\operatorname{MMD}$ estimator 
\begin{equation}
    \widehat{\operatorname{MMD}^2} \coloneqq \widehat{\lVert \mu_{\mathbb{P}_X} \rVert_{\mathcal{H}}^2} + \widehat{\lVert \mu_{\mathbb{P}_Y} \rVert_{\mathcal{H}}^2} - 2 \widehat{\langle \mu_{\mathbb{P}_X}, \mu_{\mathbb{P}_Y} \rangle_{\mathcal{H}}}
\end{equation}
and the $\operatorname{CMS}$ estimator
\begin{equation}
    \widehat{\operatorname{CMS}} \coloneqq \frac{\widehat{\langle \mu_{\mathbb{P}_X}, \mu_{\mathbb{P}_Y} \rangle_{\mathcal{H}}}}{\widehat{\lVert \mu_{\mathbb{P}_X} \rVert_{\mathcal{H}}^2} \widehat{\lVert \mu_{\mathbb{P}_Y} \rVert_{\mathcal{H}}^2}}.
\end{equation}

To reduce the runtime complexity, the $\operatorname{MMD}$ and $\operatorname{CMS}$ estimator are computed based on mini-batches of size 150 and then averaged across all blocks similar to \citep{zaremba2013b}.
We use a total of 1200 CelebA and 1350 ChestMNIST test instances for computing the errors at each epoch.
The $\operatorname{CKA}$ is computed on mini-batches of size 100, and we used 1000 CelebA training instances and 2000 ChestMNIST training instances for its computation.

\section{Practical Time and Space Complexity}
\label{app:time_space_compl}

In the following, we discuss the time and space complexity of Algorithm~\ref{alg:disent_cms}.
Let $n_{\mathrm{tr}}$ be the size of the training data, and $d = wh$ the number of pixels.
The CKA estimator of Equation~\ref{eq:CKA_est} has a runtime complexity of $O \left( n^3_{\mathrm{tr}} \right)$ and a space complexity of $O \left( n^2_{\mathrm{tr}} \right)$ due to multiplication of the kernel matrices.
In Algorithm~\ref{alg:disent_cms}, we compute the CKA estimator in a nested for-loop over the number of pixels, resulting in a runtime complexity of $O \left( d^2 n^3_{\mathrm{tr}} \right)$.
The space complexity does not increase since we do not require the respective kernel matrices after computing each CKA value.
We can reduce the runtime complexity via the following.
We may split the training data into mini-batches of size $m_{\mathrm{CKA}}$ and then average the CKA values across all mini-batches.
This results in a runtime complexity of $O \left( d^2 m_{\mathrm{CKA}}^2 n_{\mathrm{tr}} \right)$ and space complexity of $O \left( m_{\mathrm{CKA}}^2 \right)$.
Further, we may use less training data since the estimator may converge with less data than available.
In our case, we used mini-batches of size $m_{\mathrm{CKA}}=100$ and $n_{\mathrm{tr}}=1000$ training data for CelebA and $m_{\mathrm{CKA}}=100$ and $n_{\mathrm{tr}}=2000$ for ChestMNIST.
The quadratic scaling with the number of pixels can be reduced using a window of pixels as kernel inputs.
However, this was not necessary in our case.

The CMS estimator has a runtime and space complexity identical to the CKA estimator assuming $n \geq n^\prime$, where $n^\prime$ is the number of generated images per iteration.
We use the same mini-batch approach as for the CKA estimator, where $m_{\mathrm{CMS}}$ is the number of instances in a mini-batch.
We also only use a subset of size $n_{\mathrm{te}}$ of the test data.
In our experiments, we use $n_{\mathrm{te}}=1200$ and $m_{\mathrm{CMS}}=150$ for CelebA, and $n_{\mathrm{te}}=1350$ and $m_{\mathrm{CMS}}=150$ for ChestMNIST.
The runtime and space complexities of the CMS estimator with respect to the number of pixels depend on the kernel choice.
They can be neglected for the RBF and Laplacian kernel.
We assume the number of chosen clusters is rather small (for example $<10$), so we omit it as a variable.

\newpage

\section{Missing Proofs}
\label{app:proves}

In the following, we present the proof for Theorem~\ref{th:cms_disentangl}.
\begin{proof}
%\vskip-0.2in
First, note that we can disentangle the mean embedding $\mu_{\mathbb{P}_X}$ due to the assumption $\operatorname{CKA}_{k^{\otimes \lvert I \rvert},k^{\otimes \lvert I^\prime \rvert}} \left( \mathbb{P}_{X_I X_{I^\prime}} \right) = 0$ and Equivalence~\ref{eq:CKA0_implies_disentangl} via
\begin{equation}
\begin{split}
    \mu_{\mathbb{P}_X} & = \mathbb{E}_{X \sim \mathbb{P}_X} \left[ \bigotimes_{i=1}^d \phi \left( X_i \right) \right] = \mathbb{E}_{X \sim \mathbb{P}_X} \left[ \bigotimes_{I \in \mathbf{I}} \bigotimes_{i \in I} \phi \left( X_i \right) \right] \\
    \overset{\text{Assumption}}&{=} \bigotimes_{I \in \mathbf{I}} \mathbb{E}_{X \sim \mathbb{P}_X} \left[ \bigotimes_{i \in I} \phi \left( X_i \right) \right] = \bigotimes_{I \in \mathbf{I}} \mu_{\mathbb{P}_{X_{I}}}.
\label{eq:mu_disentangl}
\end{split}
\end{equation}
The same steps apply for $\mu_{\mathbb{P}_Y}$ as well.
Further, note that for any $g_1, g_2 \in \mathcal{H}$ and $h_1, h_2 \in \mathcal{H}^\prime$ we have $\left\langle g_1 \otimes h_1, g_2 \otimes h_2 \right\rangle_{\mathcal{H} \otimes \mathcal{H}^\prime} = \left\langle g_1, g_2 \right\rangle_{\mathcal{H}} \left\langle h_1, h_2 \right\rangle_{\mathcal{H}^\prime}$. Now, we can use the disentangled mean embeddings to also disentangle the overall $\operatorname{CMS}$ into a product of cluster-wise $\operatorname{CMS}$ as stated in Theorem~\ref{th:cms_disentangl}, since
%\overset{=}{\text{Eq.~\ref{eq:mu_disentangl}}}
\begin{equation}
\begin{split}
    \operatorname{CMS}_{k^{\otimes d}} \left(\mathbb{P}_X, \mathbb{P}_Y \right) & = \frac{\left\langle \mu_{\mathbb{P}_X}, \mu_{\mathbb{P}_Y} \right\rangle_{\mathcal{H}^{\otimes d}}}{\left\lVert \mu_{\mathbb{P}_X} \right\rVert_{\mathcal{H}^{\otimes d}} \left\lVert \mu_{\mathbb{P}_Y} \right\rVert_{\mathcal{H}^{\otimes d}}} = \frac{\left\langle \bigotimes_{I \in \mathbf{I}} \mu_{\mathbb{P}_{X_{I}}}, \bigotimes_{I \in \mathbf{I}} \mu_{\mathbb{P}_{Y_{I}}} \right\rangle_{\mathcal{H}^{\otimes \lvert I \rvert}}}{\left\lVert \bigotimes_{I \in \mathbf{I}} \mu_{\mathbb{P}_{X_{I}}} \right\rVert_{\mathcal{H}^{\otimes \lvert I \rvert}} \left\lVert \bigotimes_{I \in \mathbf{I}} \mu_{\mathbb{P}_{Y_{I}}} \right\rVert_{\mathcal{H}^{\otimes \lvert I \rvert}}} \\
    & = \prod_{I \in \mathbf{I}} \frac{\left\langle \mu_{\mathbb{P}_{X_{I}}}, \mu_{\mathbb{P}_{Y_{I}}} \right\rangle_{\mathcal{H}^{\otimes \lvert I \rvert}}}{\left\lVert \mu_{\mathbb{P}_{X_{I}}} \right\rVert_{\mathcal{H}^{\otimes \lvert I \rvert}} \left\lVert \mu_{\mathbb{P}_{Y_{I}}} \right\rVert_{\mathcal{H}^{\otimes \lvert I \rvert}}} = \prod_{I \in \mathbf{I}} \operatorname{CMS}_{k^{\otimes \lvert I \rvert}} \left({\mathbb{P}_{X_{I}}}, {\mathbb{P}_{Y_{I}}} \right).
\end{split}
\end{equation}    
\end{proof}

Based on the property that $\operatorname{CMS}$ always lies within $[-1, 1]$, Theorem~\ref{th:cms_disentangl} directly leads to the following fact relevant for interpretation.
\begin{corollary}
    Under the same assumptions as in Theorem~\ref{th:cms_disentangl}, it holds for all $I \in \mathbf{I}$ that
    \begin{equation}
        \left\lvert \operatorname{CMS}_{k^{\otimes d}} \left( \mathbb{P}_X, \mathbb{P}_Y \right) \right\rvert \leq \left\lvert \operatorname{CMS}_{k^{\otimes \lvert I \rvert}} \left( \mathbb{P}_{X_{I}}, \mathbb{P}_{Y_{I}} \right) \right\rvert.
    \end{equation}
\end{corollary}
In other words, the $\operatorname{CMS}$ of the whole image grid can never surpass the $\operatorname{CMS}$ of any cluster.
This important fact tells us that we may never expect a smaller similarity between prediction and target in any cluster compared to the overall similarity.
If this property is violated in practice, we will have to be wary of violated assumptions, which may affect the correctness of interpretations based on Theorem~\ref{th:cms_disentangl}.

%%%%%%%%%%%%%%%%%%%%%%%%%%%%%%%%%%%%%%%%%%%%%%%%%%%%%%%%%%%%

\end{document}